\documentclass[twocolumn,english]{IEEEtran}
\usepackage[T1]{fontenc}
\UseRawInputEncoding
\usepackage{array}
\usepackage{verbatim}
\usepackage{graphicx}

\makeatletter

\usepackage[hidelinks]{hyperref}
\usepackage[capitalise,nameinlink]{cleveref} 

\usepackage{authblk}            
\usepackage{orcidlink}          

\newcommand{\noun}[1]{\textsc{#1}}
\providecommand{\tabularnewline}{\\}

\makeatother

\usepackage{babel}

\begin{document}
\title{HW/SW Co-design of a PCM/PWM converter: a System Level Approach based in the SpecC Methodology}


\author{Daniel G. P. Petrini\,\orcidlink{0000-0002-7278-9632}\thanks{E-mail: \href{mailto:dpetrini@alumni.usp.br}{dpetrini@alumni.usp.br}}
\and Braz Izaias da Silva Junior}

\maketitle

\thispagestyle{empty}

\begin{abstract}
[Original work from 2005; formatting revised in 2025, with no changes to the results.] 
We present a case study applying the SpecC methodology within a system-level hardware/software co-design flow to a PCM-to-PWM converter, the core of a Class-D audio amplifier. The converter was modeled and explored with SpecC methodology to derive an HW/SW partition. Using system-level estimates and fast functional simulation, we evaluated mappings that meet real-time constraints while reducing estimated cost of an all-hardware solution and avoiding the expense of a purely software implementation on a high-end processor. Despite the design's moderate complexity, the results underline the value of system-level co-design for early architectural insight, rapid validation, and actionable cost/performance trade-offs.
\end{abstract}


\section{Introduction}

\pagestyle{empty}


The recent requirements of the semiconductors industry
has lead the design methodologies evolution in order to fill the design
gap detected by the 1999 Roadmap \cite{EITR:1999}. Today design
has very high integration density and complex functionalities to implement
arising the necessity to use a higher level of abstraction, the so-called
System Level. The latter has many approaches like: reuse methodologies
\cite{RMM:1999}, Intellectual Property cores, platform-based-design
\cite{PLB:2001}, specification and design languages for the system
level and hardware/software co-design. 
In the latter, functional partitioning and mapping to hardware or software are analyzed to optimize objectives such as cost and time-to-market.

In this paper we present a case study of a design made in the System
Level approach using the system level design language SpecC to obtain
a hardware and software partitioning. In the next section we mention
the design phases of co-design. In section III, co-design computer
aided tools are analyzed and in section IV, SpecC is reviewed. Section
V contains the description of the case study followed by its results
and conclusion.

\section{Related Work}

In a hardware/software co-design environment, the project can be
divided in phases \cite{KAL:1993}. These phases are specification,
modeling, representation, estimation, partitioning, co-synthesis,
co-simulation and eventually co-verification. 
This study focuses primarily on the estimation and partitioning phases of the design flow.

An estimator is a tool that, when applied to a candidate implementation, produces quantitative predictions for relevant metrics, thereby enabling comparative evaluation of design alternatives \cite{GAJ:1995}.

Partitioning may proceed in a software-oriented manner-starting from a 100\% software implementation and progressively migrating selected functional blocks to hardware \cite{COS:1993}-or in the reverse direction, beginning with a hardware-based solution and incrementally mapping functions to software \cite{VUL:1992}. The process first \emph{divides the system} into groups or clusters at an appropriate level of granularity; these resulting blocks are then \emph{mapped to processing elements} (the physical components). Selecting the specific components constitutes \emph{architectural allocation}. From there, blocks are assigned to hardware or software according to an \emph{objective (cost) function} that guides the trade-offs among design goals \cite{COS:1993,VAL:2000}.

Co-design methodologies and tools differ in how they produce a hardware/software partition-either automatically or manually (designer-driven). \emph{Automatic partitioning} integrates algorithms within the tool that ingest the captured system model and propose an optimal split according to a defined cost function, optionally with limited designer guidance. Common algorithmic families include \emph{constructive} methods \cite{VAL:2000,REC:2002} and \emph{interactive} approaches \cite{VUL:1992}.
In \emph{manual partitioning}-typically the outcome of design space exploration-functional logic blocks are grouped by affinity as expressed in a high-level design-language specification. The identification of these groups (often termed behaviors) is performed by the designer, guided by the architectural and logical characteristics of the system, as in \cite{JPE:2000}.

\section{Co-design Tools}

\label{sec:Co-design-Tools}




The primary purpose of design tools is to automate and accelerate key phases of the development flow, enabling modeling, simulation, validation, and synthesis from a unified system representation.

A wide range of models of computation underpin these tools, including finite-state machines (FSMs), extended FSMs \cite{CHI:1994}, data-flow graphs, statecharts, data-path FSMs \cite{GAJ:1991}, program FSMs (the basis of SpecC) \cite{GER:2001} and Petri nets. This subject is extensively discussed in the Ptolemy project \cite{PTP:2001}.

Within co-design, the specification should capture the system's full functionality. This specification serves as an executable model that encodes system properties and supports simulation. It forms the foundation for system-level design exploration-extracting, analyzing, and experimenting with alternative architectural solutions.

Our survey of the area reveals two main classes of co-design tools. The first comprises toolchains originally oriented toward \emph{board-level design}-i.e., printed-circuit boards integrating a CPU, memories, and an ASIC or FPGA-which later converged toward to a second approach that aims towards to the \emph{system-on-chip (SoC)} paradigm, where these elements are integrated into a single device. Representative tools in the first group include Cosyma \cite{COS:1993}, Vulcan II \cite{VUL:1992}, Ptolemy \cite{PTP:2001}, PISH \cite{PIH:1997}, Polis \cite{POL:1994}, and SpecSyn \cite{GAJ:1998}. Tools emblematic of the second, SoC-focused group include CoWare \cite{ECOW:2004}, SpecC \cite{GER:2001}, SystemC \cite{ESYS:2004}, and, more recently, Spark, which raises the abstraction level by compiling high-level descriptions into HDL \cite{ESPA:2003}.

For the case study reported in Section \ref{sec:Case-Study--}, we compared several of these tools using the following criteria: (a) compatibility of the modeling formalism with our specification (a C program), (b) support for design-space exploration of the hardware/software partition-preferably with automated assistance, (c) quality of documentation, and (d) availability. Based on this evaluation, we selected a tool grounded in the SpecC language and methodology: the \emph{System-on-Chip Environment (SCE)}, alpha version, which was made available to us by its creators at the University of California, Irvine.

\section{SpecC Methodology}

\begin{figure}[hbt]
\begin{centering}
\includegraphics[width=0.95\columnwidth,keepaspectratio]{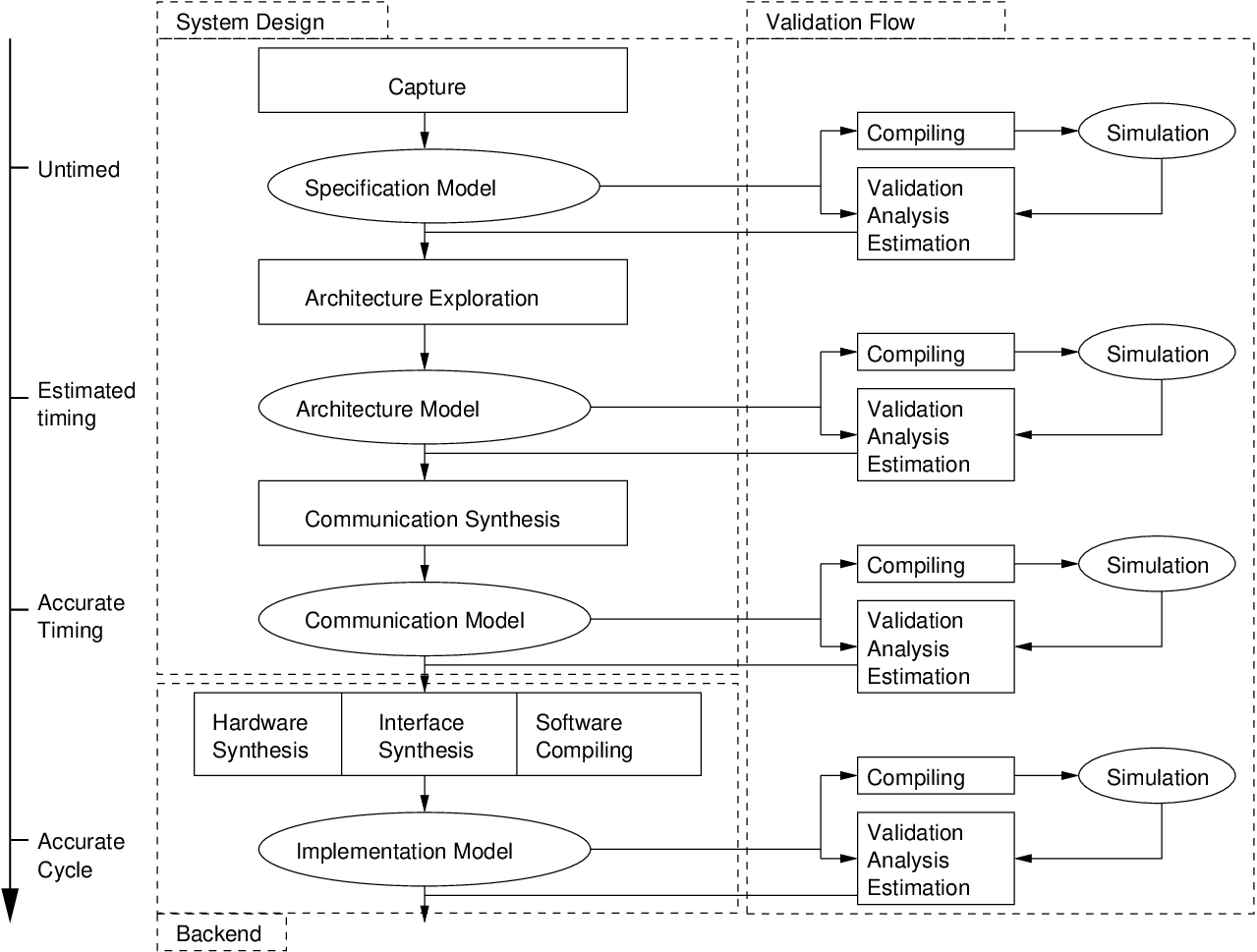}
\par\end{centering}
\caption{SpecC Models.}

\label{cap:SpecC-Models.}
\end{figure}

The SpecC methodology is a tool developed to support high levels of
abstraction designs. It aims to refine an abstract design specification
into a implementation model running in the target architecture. It
is based in four models, namely a specification model, an architecture
model, a communication model and finally an implementation model.
In figure \ref{cap:SpecC-Models.} we see these models and the three
transformations, architecture Exploration, Communication Synthesis
and backend, which comprises hardware and interface synthesis and
software compiling, as in \cite{GER:2001}.

The first tree models are written in SpecC language. Each one is compiled
and simulated until its validation. Estimations are obtained by operations
counting during simulation together with information of the processing
elements delays per operation. 

The \emph{Specification Model} is a functional description of the system. To create this model, the requirements must be analyzed and translated into the SpecC language. The system is then divided into blocks (behaviors) with similar functionality, thereby defining the granularity and revealing potential parallelism. Simulation at this stage is purely functional (zero-delay), with no timing considerations.

The \emph{Architecture Model} is obtained after \emph{architecture exploration} and \emph{architecture refinement}. Architecture exploration evaluates design alternatives, using behavior estimates and electronic component characteristics to produce a manual hardware/software partition. Architecture refinement then applies transformations such as behavior and variable optimizations. The resulting model reflects the impact of the mapped computation on the chosen architecture and captures the desired hardware-software division.

\emph{Communication Model} is reached after communication refinement, which
adds system buses to the design and maps the abstract communication
between components onto the busses. This refinements includes tasks
such as communication channels partitioning, protocol insertion above
the abstraction of channel and protocol in-lining. The result is a
model with much more timing accuracy.

The \emph{Implementation Model} is the model with the lowest level of abstraction. The SpecC code from the previous model, related to the hardware behaviors, should be translated to a language accepted by high-level synthesis as VHDL, because SpecC is not yet synthesizable. The software behaviors should
be translated to standard C++ language to be compiled to the target
processor. The interface is already allocated in the software or hardware
behaviors, so should not have different handling.

\begin{figure}[hbt]
\begin{centering}
\includegraphics[width=0.7\columnwidth]{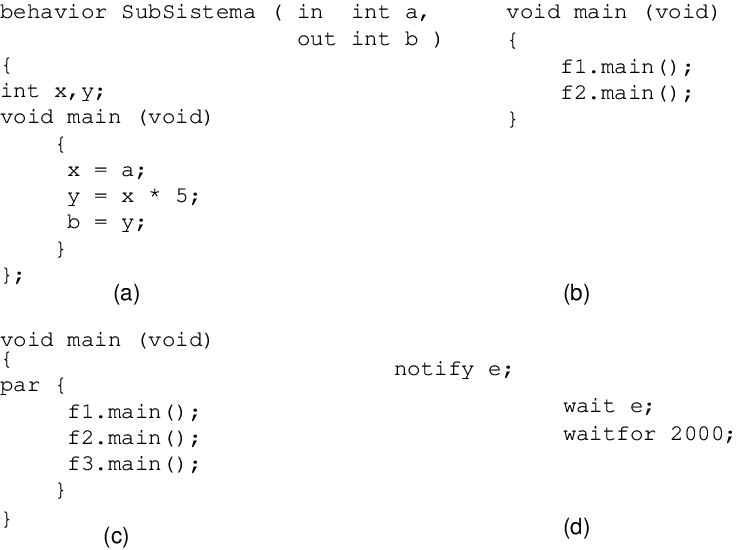}
\par\end{centering}
\caption{SpecC Constructions: (a)Behavior declaring, (b)Sequential execution,
(c)Parallel execution, (d)Synchronization elements.}

\label{cap:SpecC-Construction:-(a)Behacior}
\end{figure}

SpecC language is a superset of C language \cite{LRM:2002}. The added
constructions allow concurrency modeling, hierarchy, communication,
synchronization, state transitions, exception handling and timing
- all necessary to embedded systems design. it uses the SpecC Reference
Compiler - \emph{scrc}. In the figure \ref{cap:SpecC-Construction:-(a)Behacior}
we can see some of its constructions.

The SCE provides a environment for modeling, synthesis and validation.
It includes a graphical user interface and the tools to facilitate
the design flow and perform the aforementioned refinements of the
SpecC methodology \cite{SCE:2003}.

\section{Case Study - The PCM/PWM converter}

\label{sec:Case-Study--}

To assess the benefits of a system-level approach for a design of intermediate complexity, we conducted a case study in which an already implemented design was re-evaluated using a hardware/software co-design methodology. The target design is an audio PCM-to-PWM converter, originally written in Verilog HDL and implemented on an FPGA

\begin{figure}[hbt]
\begin{centering}
\includegraphics[width=0.95\columnwidth]{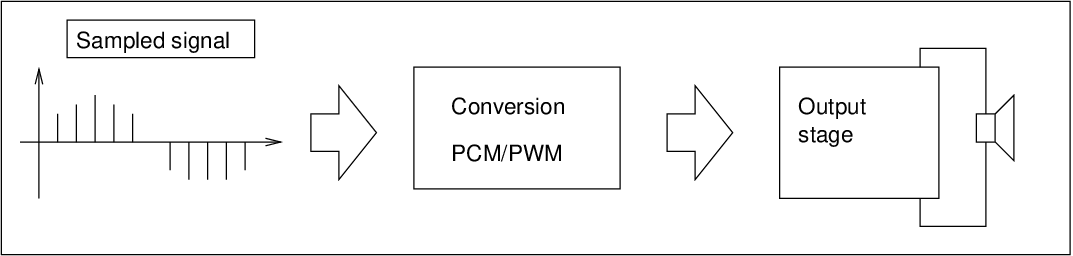}
\par\end{centering}
\caption{Class D digital amplification. }

\label{cap:Class-D-digital}
\end{figure}

Most audio power amplifiers operate in Class A or Class AB configurations. In these cases, the amplifier dissipates too much power, resulting in lower efficiency. By contrast, in digital (switching) amplification, power dissipation is reduced. This category, known as Class-D digital amplifiers, can achieve efficiencies of around 90\%.
This flow can be seen in figure \ref{cap:Class-D-digital}.

As the conversion itself is not the main focus of this study, so is
the design methodology, we will briefly describe its working principle
and its main blocks. 

In the pulse-width modulation (PWM), the pulse amplitude is kept constant
while the width is proportional to signal samples. In a rough approximation,
this pulse width should have the same resolution of the samples, this
means that if the signal samples is quantized in $n$ levels, the
pulse width should be quantized in $n$ levels as well. For a CD audio
signal we have $2^{16}$quantization levels and the sampling frequency
is 44,1 kHz. To transform the amplitude in pulse width we would need
a $2^{16}x44100Hz=2,89GHz$ resolution. Such frequency is not suitable
for implementation in cost expectations of consumer electronics. To
solve this problem, we need to add signal processing techniques to
convert the PCM into PWM. The main goal is to decrease the operational
frequency. Accordingly, we developed a technique that makes this feasible, comprising an algorithm divided into four stages:
(a) an up-sampling block that increases the sampling rate by digital
interpolation; (b) as PW modulation inserts delays causing harmonic
distortion, the linearization block compensated this; (c) the noise
shaper block which adjusts the trade-off between number of quantization
levels and the new sampling rate; and (d) a wave form generation which
performs the pulse creation according to output stage needs.

\begin{figure}[hbt]
\begin{centering}
\includegraphics[width=0.95\columnwidth]{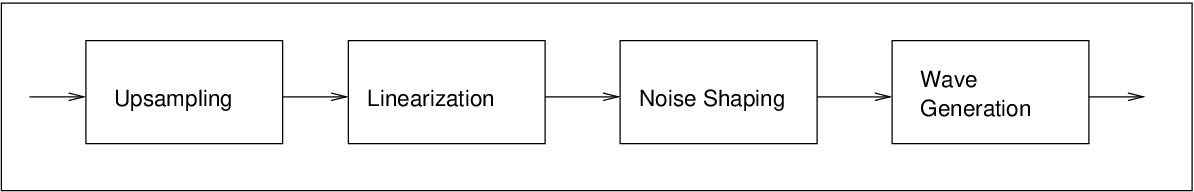}
\par\end{centering}
\caption{PCM-PWM conversion stages, which guide the subsequent behavior mapping: S0, S1, S2, and S3 (upsampling), LINE (linearization), and MOLD (noise shaping).}

\label{cap:PCM/PWM-conversion-phases.}
\end{figure}

This stages happens sequentially as shown in figure \ref{cap:PCM/PWM-conversion-phases.}. The frequency reached is about 45 MHz, in its original implementation, which is suitable for consumer electronics applications.

\section{SpecC Usage in case study}

The algorithm development were done in Matlab environment and then
translated to IBM-PC platform software in order to be validated. Once
this algorithm was working in PC, it was converted manually to Verilog
hardware description language and implemented in FPGA. Our goal is,
from the same specification, which was the C program running in PC
platform, develop the co-design case study using SpecC methodology.
The implementation in FPGA reached the timing requirements but it
ended up requiring a high number of programmable logic elements, so reached a
high cost. 

The first task was to convert the C program into a SpecC program. In
order to validate this process, we developed a testbench that consists of a file containing about four seconds of audio. These
samples were converted to PWM by the SpecC model and saved in a file.
This file was then reconverted to audio by a command line tool which
applied the opposite algorithm and the results were compared to initial
audio file by subjective audition.

SCE provides a library of processing elements (PEs) along with associated statistics. During component allocation, the most suitable PEs are selected from this library. Because we used an alpha version, the library was neither complete nor fully accurate; however, it contained sufficient information for our purposes. Table \ref{cap:Features-ofthe-components} lists the component characteristics, including cost (in U.S. dollars). The sole exception is the hardware cost, which was derived from the original design's FPGA implementation rather than from the SCE library.

\begin{table}[hbt]
\begin{centering}
\begin{tabular}{|l|l|l|c|r|}
\hline 
Element & Description & Category & Cost & F(MHz)\tabularnewline
\hline 
\hline 
DSP & DSP 56600 Motorola & DSP & 8 & 60\tabularnewline
\hline 
uP & Intel Pentium III & Microproc. & 40 & 900\tabularnewline
\hline 
uC & Siemens 80166 & Microcontrol. & 1 & 8\tabularnewline
\hline 
HW & Standard RTL processor & FPGA & 35 & 100\tabularnewline
\hline 
\end{tabular}
\par\end{centering}
\caption{Characteristics of the components. }

\label{cap:Features-ofthe-components}
\end{table}

\subsection{Specification Model}

\begin{figure}[hbt]
\begin{centering}
\includegraphics[width=0.8\columnwidth]{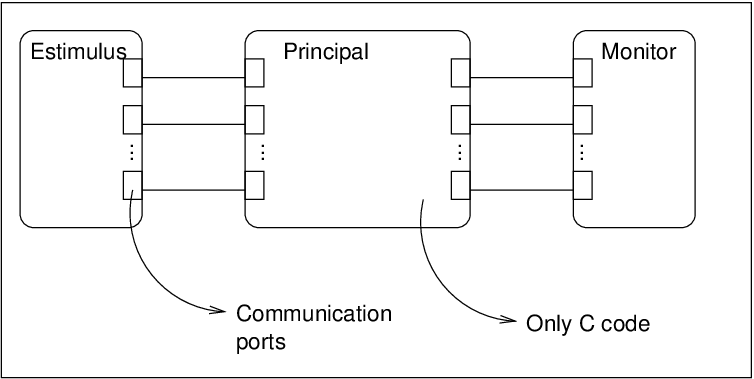}
\par\end{centering}
\caption{Connectivity between behaviors in the testbench for phase \# 1 towards
specification model.}

\label{cap:Conectivity-between-behaviors}
\end{figure}

Use of the SpecC tool begins with constructing the \emph{Specification Model}. Our first step was a straightforward translation from C to SpecC, wrapping the original C code in SpecC constructs to establish the testbench. At this stage, the entire code resides in a single SpecC behavior (behavior \emph{Principal} in Fig. \ref{cap:Conectivity-between-behaviors}). Although the design can already be simulated and validated in the SCE environment, further refinement is required to divide this behavior into more partitions.

The next step was to separate convenient groups of functionalities into individual behaviors to facilitate analysis. The computations were grouped into three behaviors corresponding to the up-sampling, linearization, and noise-shaping phases. In the SpecC code, the original C segments implementing these functions were replaced by calls to the newly defined behaviors, yielding code that closely resembles a C program with standard function calls. The control structure, four nested `for` loops, was left unchanged at this point.

\begin{figure}[hbt]
\begin{centering}
\includegraphics[width=0.8\columnwidth,keepaspectratio]{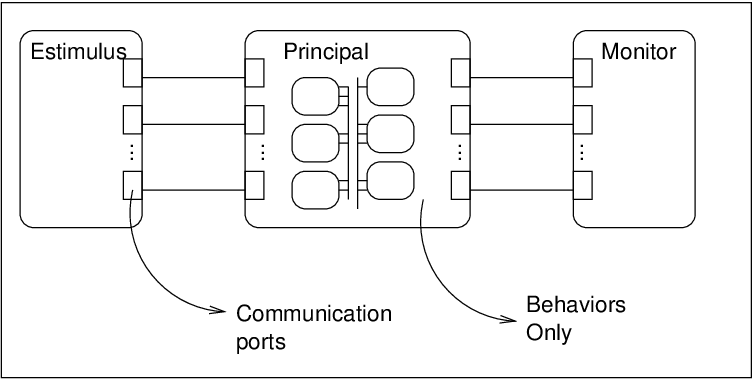}
\par\end{centering}
\caption{Connectivity of top-level behavior Principal, now only containing
leaf-behaviors inside.}

\label{cap:Conectivity-of-top-level-Phase_3}
\end{figure}

The third and final step was more challenging, as it required a modeling style compatible with SCE's automatic tooling for subsequent design phases. To achieve this, the program structure was refactored into a program finite-state machine (program FSM), with each state representing a distinct program-FSM stage. Figure \ref{cap:Conectivity-of-top-level-Phase_3} shows the resulting system, which contains only leaf behaviors within the top-level behavior \emph{Principal}. It is worth noting that this re-modeling of the algorithm's control flow took longer than anticipated to reach the specification model.

The final specification model was composed by six state-behaviors
that runs on a fsm fashion. Our intention was to use the minimum number
of behaviors because for each added behavior there is a overhead of
communication, due to the use of communication ports.

\subsection{Architecture Model}

The first step toward this model is \emph{architecture exploration}, a key advantage of our system-level approach, as it quickly reveals how the system should be organized.

Architecture exploration begins by gathering information about the computational cost of the system across different processing elements. In SpecC, estimates are derived from the number of operations counted during simulation. Given this count, we compute execution times for each target platform by applying \emph{weighted operation costs}. For example, on a DSP, a multiply-accumulate (MAC) may require a single cycle, whereas on a general-purpose processor it may take multiple cycles. Combining the weighted operation counts with each component's operating frequency yields the estimated execution time for the design, as shown in Table \ref{cap:Principal-behavior-estimation}.

\begin{table}[hbt]
\begin{centering}
\begin{tabular}{|l|c|c|c|c|}
\hline 
Element & Type & Cycles & $\Delta$t Execution(s) & Goal\tabularnewline
\hline 
\hline 
DSP & SW & 272.935.511 & 4,54 & \tabularnewline
\hline 
uP & SW & 935.152.757 & 1,04 & X\tabularnewline
\hline 
uC & SW & 935.152.757 & 116,89 & \tabularnewline
\hline 
HW & HW & 250.224.089 & 2,50 & X\tabularnewline
\hline 
\end{tabular}
\par\end{centering}
\caption{Principal behavior estimation in all components.}

\label{cap:Principal-behavior-estimation}
\end{table}

\begin{figure}[hbt]
\begin{centering}
\includegraphics[width=0.8\columnwidth,keepaspectratio]{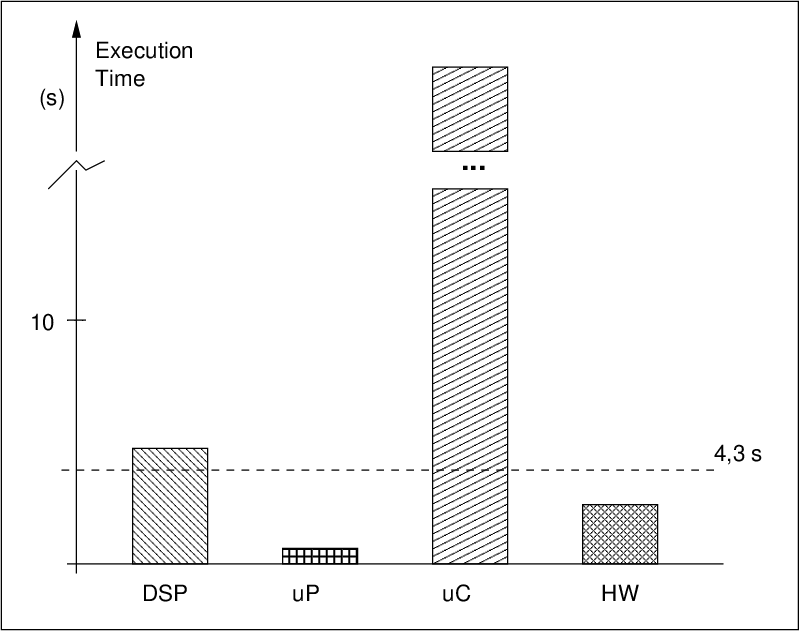}
\par\end{centering}
\caption{Execution time for different target candidates and the threshold time.}

\label{cap:Execution-time-for}
\end{figure}

The objective is met when the execution time is shorter than the input file's playback duration, that is, when real-time performance is achieved. We can see it
graphically in figure \ref{cap:Execution-time-for}. SCE profiler
gives other kind of estimates, such as amount of communication, in
bytes, between behaviors, variable counting and others (not shown here).

From the estimated execution times, only the Pentium processor (uP) and the fully hardware-based implementation satisfy the real-time requirement. The Pentium option, being a purely software solution that departs from the co-design premise, is also cost-prohibitive and is therefore discarded. Although a full-hardware solution is expensive, we plan to use it only partially, so we continue its analysis. We do the same for the DSP, which nearly met the target and offers an attractive cost profile. The microcontroller (uC), another 100\% software option, is eliminated due to inadequate performance for this type of system.

To advance the analysis, we introduce the implementation cost of each behavior, introduced in Figure \ref{cap:PCM/PWM-conversion-phases.}, on each selected processing element (PE). Using the \emph{code size} metric available in SCE (beta) for each PE, we compute the per-behavior cost proportional to its code size and to the PEs total cost. For the \emph{HW} and \emph{DSP} PEs, the base costs are taken from Table \ref{cap:Features-ofthe-components}; for the DSP, we add one cost unit for memory, yielding a fixed cost of 9,00. Our intent is to employ both the DSP and HW components. The DSP's cost is fixed, independent of the number of mapped behaviors, since it is entirely software. By contrast, the hardware cost varies with the number of mapped behaviors, the less the number of behavior, the smaller the FPGA needed. Table \ref{cap:Execution-time-(in} reports per-behavior execution times for the two processing elements, FPGA (hardware) and DSP (software), along with the corresponding proportional costs for the FPGA.

\begin{table}[hbt]
\begin{centering}
\begin{tabular}{|p{9,5mm}|p{4mm}|p{5mm}|p{6mm}|p{6mm}|p{5,5mm}|>{\centering}p{7mm}|>{\centering}p{7mm}|}
\hline 
 & S0 & S1 & S2 & S3 & LINE & MOLD & Total\tabularnewline
\hline 
\hline 
$\Delta$t HW & 2,2 & 183,3 & 502,0 & 988,1 & 77,4 & 749,1 & 2502,5s\tabularnewline
\hline 
(\$) HW & 0,01 & 8,17 & 10,73 & 10,65 & 3,60 & 1,60 & \$35,00\tabularnewline
\hline 
$\Delta$t DSP & 5,4 & 305,6 & 836,7 & 1646,4 & 188,1 & 1566,7 & 4548,9s\tabularnewline
\hline 
\end{tabular}
\par\end{centering}
\caption{Execution time (ms) and cost (US\$). Note that all behaviors mapped to software (DSP) sum to more than 4.5 s, thereby failing to meet the 4.3 s real-time requirement.}

\label{cap:Execution-time-(in}
\end{table}

An analysis of possibilities, or \emph{architecture mapping}, is carried
out with the results in the Table \ref{cap:Execution-time-(in}. We analyze the trade-off
between required timings, costs and available resources. Our goal
is to maximize the software usage because that reflects in cost reduction
in this design.

\begin{table}[hbt]
\begin{centering}
\begin{tabular}{|c|c|c|c|c|c|}
\hline 
 & Mapped in HW & $\Delta$t DSP & $\Delta$t HW & $\Delta$t total & Total cost\tabularnewline
\hline 
\hline 
1 & S3 & 2902,5 & 988,1 & 3890,6 & 19,65\tabularnewline
\hline 
2 & S1, S2, S3 & 1760,2 & 1673,4 & 3433,6 & 38,55\tabularnewline
\hline 
3 & LINE,MOLD & 2794,1 & 826,5 & 3620,6 & 14,20\tabularnewline
\hline 
4 & MOLD & 2982,0 & 749,1 & 3731,0 & 10,60\tabularnewline
\hline 
\end{tabular}
\par\end{centering}
\caption{Execution time (ms) and cost analysis varying the set of behaviors mapped in HW.}

\label{cap:Execution-time-(ms)}
\end{table}

From Table \ref{cap:Execution-time-(ms)}, \emph{Option 1} maps the heaviest behavior (\emph{S3}) to hardware, meeting the timing constraint at a reasonable cost. However, \emph{S1}, \emph{S2}, and \emph{S3} share the same FIR filter, so placing all three on a single PE is attractive; \emph{Option 2} evaluates this and attains very aggressive timing, albeit at high cost. \emph{Option 3} maps \emph{LINE} and \emph{MOLD} to hardware due to their affinity (shared communication buffers). \emph{Option 4} maps only \emph{MOLD} to hardware and yields the lowest implementation cost.
Comparing \emph{Options 3 and 4}, adding \emph{LINE} (Option 3) improves timing by only \emph{3.05\%} but increases cost by \emph{25.53\%}. Since both total times (3620.6 ms and 3731.0 ms) are below the 4300 ms threshold, \emph{cost} is the decisive metric, favoring \emph{Option 4}.

These were the most relevant execution times and costs to consider. Because the system must run sequentially, options that assume non-sequential behaviors are unsuitable. \emph{State S0} was mapped to the DSP due to its small size and proximity to \emph{S1}. The PWM waveform generator, the final block in Fig. \ref{cap:PCM/PWM-conversion-phases.}, is treated separately and can be mapped either to hardware or to software. Implemented in hardware, it offers high performance with minimal area/cost, as it consists only of a counter, a register, and a comparator. Alternatively, on a software PE it can be realized via an internal peripheral (e.g., the \emph{DSP56003/005} include five on-chip PWMs). For these reasons, it was excluded from the exploration calculations.

\begin{figure}[hbt]
\begin{centering}
\includegraphics[width=0.8\columnwidth,keepaspectratio]{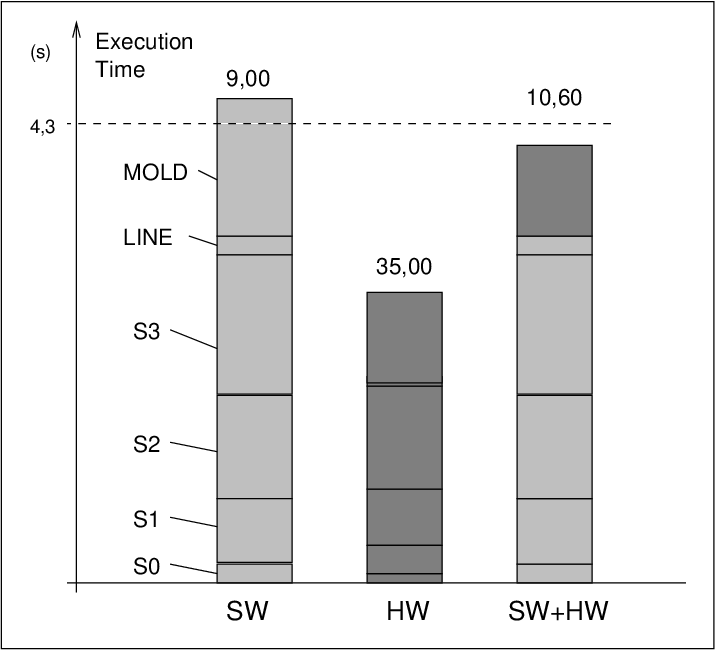}
\par\end{centering}
\caption{Co-design approach result. Components cost above the bars.}

\label{cap:Comparison-between-architectures.}
\end{figure}

According to the analysis in paragraphs above, mapping \emph{MOLD} to hardware satisfies the timing constraints while reducing cost relative to an all-hardware implementation. Figure~\ref{cap:Comparison-between-architectures.} shows execution time and per-behavior cost for three alternatives: a \emph{100\% software} approach on the DSP, a \emph{100\% hardware} solution, and a \emph{co-design} configuration that meets the computational requirements at an attractive cost. Thus, we manually defined the system's hardware/software partition.

Once concluded the design space exploration, SCE tool performs automatically
the architecture refinement transformation optimizing variables (that
serve as communication between behavior mapped in different PEs) and
scheduling of behaviors. Thus, we reach architecture model and the
system is validated by simulation as in figure \ref{cap:SpecC-Models.}.

\subsection{Communication and Implementation models}

In the communication model, a bus is chosen and its timing is considered
in timing analysis. The \emph{communication refinement} automatic
transformation, inserts handshake protocol and other details to reflect
the communication between PEs and bus. We then validate the system via simulation, which takes longer than before due to the added detail.

The resulting communication model serves as input to the implementation (backend) phase. At this stage, SpecC modules designated as software must be translated into C for the target DSP, while hardware behaviors should be converted into HDL and synthesized. Because the SCE version used did not fully support these features, we did not perform this step in our case study.

\section{Results}

For our case study, the SpecC/SCE methodology offered several advantages: (a) the use of estimates to inform architectural decisions; (b) automatic refinement steps that shorten development time; (c) fast, system-level simulation—less precise but much quicker, enabling earlier validation; and (d) feasible manual exploration thanks to a small number of system behaviors. We also observed limitations: (a) absence of automatic partitioning algorithms; (b) a limited set of exploration metrics; and (c) a reduced processing-element library. It is important to note that the evaluated version was a work in progress, and these gaps may be addressed in future releases.

We also defined comparison criteria between the original RTL-based design flow and the system-level co-design flow used in the case study. Focusing on the most relevant factors for low cost and time-to-market, the summary appears in Table \ref{cap:Comparison-between-co-design}.

\begin{table}[hbt]
\begin{centering}
\begin{tabular}{|c|p{1in}|p{1in}|}
\hline 
Criterion & Co-design & RTL\tabularnewline
\hline 
\hline 
Remodeling effort & \multicolumn{1}{>{\centering}p{1in}|}{High to control flow/ low to signal processing} & High to both control flow and signal processing\tabularnewline
\hline 
Simulation time & Specif. Model: 1min, Comm.Model: 30 min & High: some hours\tabularnewline
\hline 
Implementation cost & Low cost DSP with reduced area for HW ("small FPGA") & High, only HW\tabularnewline
\hline 
Reuse & Not used, but SpecC supports & Yes, some functions were reused\tabularnewline
\hline 
Lines of Code & 800 lines SpecC/C - 500 added & Verilog \textasciitilde{} 10k lines\tabularnewline
\hline 
Refinement & Low effort, Automatic & High effort, manual convertion\tabularnewline
\hline 
HW/SW Partitioning & Manual with only 6 behaviors & No partitioning, only HW\tabularnewline
\hline 
\end{tabular}
\par\end{centering}
\caption{Comparison between Hardware/Software co-design and RTL.}

\label{cap:Comparison-between-co-design}
\end{table}

The conversion of C code to SpecC code to reach specification model
and the architecture exploration were the phases that dispensed most
time of the work. The effort to remodel the code were reasonable, even
dealing in the same abstraction level (including the fact that the experience
with SpecC grew during the study), but as mentioned in the table above,
the computation internal to behaviors were not modified in the C to
SpecC conversion.

In the design space exploration phase, we used three criteria, execution
time, monetary cost and designer experience. The first two were provided
by the project environment and the last is experience from past designs.
In our case study the first two were enough to the result.

\section{Conclusions}

Although we did not produce a final implementation, the evidence summarized in Table \ref{cap:Comparison-between-co-design} indicates that the adopted approach was satisfactory, yielding meaningful cost optimization. The PCM/PWM converter presents moderate computational complexity, making it amenable to both the co-design flow used here and a conventional RTL flow. Nonetheless, as system complexity continues to grow, system-level co-design becomes increasingly recommended.
The signal-processing nature of the algorithm aligned well with SpecC, whose C-based foundation supports broad applicability. For problems specified directly in high-level languages, or easily translated to them, system-level methodologies such as SpecC are effective for producing implementation proposals with actionable estimates, while accelerating development cycles through rapid simulation and early validation.

~

\noun{Acknowledgements}: We would like to thank Rainer D{\"o}mer from
University of California, Irvine.

\bibliographystyle{ieeetr}
\bibliography{dpetrini_references}

\end{document}